\documentclass[11pt]{article}

\usepackage[]{acl}

\usepackage{times}
\usepackage{latexsym}
\usepackage{amsmath}
\usepackage{booktabs}
\usepackage{multirow} 
\usepackage{xcolor}

\usepackage[T1]{fontenc}

\usepackage[utf8]{inputenc}

\usepackage{microtype}

\usepackage{inconsolata}

\usepackage{graphicx}

\title{Diagnosing and Mitigating Thinking Collapse in On-Policy Self-Distillation}

\author{%
  Keqin Peng$^{1}$,
  Chen Li$^{1}$,
  Yuanxin Ouyang$^{1}$,
  \textbf{Yancheng Yuan}$^{2*}$,
  \textbf{Liang Ding}$^{3}$\thanks{~~Corresponding Authors.}\\
  $^{1}$Beihang University $^{2}$Hong Kong Polytechnic University $^{3}$Alibaba Group\\
\texttt{keqin.peng@buaa.edu.cn}\quad\texttt{liangding.liam@gmail.com}}

\begin{document}
\maketitle
\begin{abstract}
On-Policy Self-Distillation (OPSD) has emerged as a crucial paradigm for enhancing and aligning Large Language Models (LLMs). However, in complex reasoning tasks, OPSD paradoxically degrades downstream performance. In this paper, we systematically investigate this pathology and identify a severe optimization trap we define as \textbf{Thinking Collapse}---a sharp decline in the model's native intermediate reasoning behavior, measured by epistemic-token density (ET per 1k). Through entropy-based gradient masking and token-level target analysis, we show that this collapse is triggered by aggressive teacher gradients at high-student-entropy decision forks, where student epistemic tokens are frequently suppressed into teacher non-epistemic targets and are highly concentrated in high pointwise student-teacher divergence regions. To resolve this optimization pathology, we propose \textbf{Adaptive Dual-Perspective OPSD (AD-OPSD)}, a robust control framework that dynamically moderates the self-distillation objective. AD-OPSD selectively anchors high-suppression-risk sandboxed tokens to a reference prior derived from the frozen base model via an asymmetrical pointwise divergence gate, preserving native thinking capacity while retaining OPSD's error-correcting power. Extensive experiments across competitive mathematical benchmarks show that AD-OPSD improves over standard OPSD by up to \textbf{+4.1\%} absolute average accuracy across diverse model scales and datasets. Further analysis demonstrates that AD-OPSD mitigates thinking collapse and generalizes robustly to different post-training paradigms.
\end{abstract}

\section{Introduction}
\label{sec:intro}

Recently, on-policy self-distillation (OPSD) has emerged as a crucial paradigm for enhancing and aligning Large Language Models~\citep{zhao2026self,shenfeld2026self,hubotter2026reinforcement}. By distilling ground-truth-conditioned, on-policy target distributions of a teacher model into the student policy over its own sampled rollouts, OPSD avoids the exposure bias of off-policy imitation and the sparse-reward bottleneck of reinforcement learning. This dense, token-level supervision has yielded remarkable success in general alignment and preference-tuning tasks.

However, when applied to complex reasoning tasks, OPSD paradoxically degrades downstream performance~\citep{kim2026does,kaur2026rethinking}. Prior studies attribute this degradation to the suppression of epistemic verbalizations~\citep{kim2026does} or related ``fork suppression'' at high-entropy decision points~\citep{kaur2026rethinking}. Yet, these works primarily document empirical symptoms; the underlying mechanics of \textit{where} and \textit{why} this suppression occurs remain unclear, leaving how to resolve this optimization trap without losing OPSD's error-correcting power an open challenge.

In this paper, we systematically investigate this pathology, identifying a severe optimization trap we define as \textbf{Thinking Collapse}---a precipitous decline in the model's native intermediate reasoning steps, measured as epistemic tokens per 1K generated tokens (\textit{ET per 1k}). To understand its mechanics, inspired by the findings of~\citet{wang2026beyond} and~\citet{xu2026tip} that reasoning updates and critical decision forks are heavily concentrated on small sets of high-entropy tokens, we propose an entropy-based gradient masking diagnostic experiment to locate the spatial boundary of thinking collapse. 
We find that masking gradients on high-student-entropy tokens substantially recovers thinking density, but also exposes a correction trade-off. Specifically, masking the self-distillation gradients of only the top 20\% student entropy tokens recovers the model's native thinking density from OPSD's collapsed \textbf{7.9} to \textbf{9.8 ET per 1k}, while still failing to recover downstream accuracy.

Furthermore, we conduct a microscopic token-level discrepancy transition analysis at initialization, discovering a substantial top-1 target discrepancy where the student's top-1 predicted epistemic token is forcefully suppressed by a teacher non-epistemic token. Lastly, mapping the student's epistemic space under active suppression ($\log P_s > \log P_t$) reveals that native thinking steps are densely concentrated in high pointwise divergence regions, supporting the use of pointwise student-teacher divergence as a robust indicator of suppression risk.

To resolve this optimization pathology, we propose \textbf{Adaptive Dual-Perspective OPSD (AD-OPSD)}, a robust control framework that dynamically moderates the self-distillation objective. Instead of binary token masking, AD-OPSD employs a continuous, asymmetrical soft-gating mechanism. Within a localized high-entropy sandbox, it selectively anchors high-suppression-risk tokens to a reference prior derived from the frozen base model using a sigmoid pointwise KL gate. This dynamically shields the model's native exploratory reasoning paths under active suppression while allowing standard teacher corrective gradients on factual tokens to proceed at full strength.

Extensive experiments across competitive benchmarks demonstrate that AD-OPSD consistently recovers thinking density and delivers up to a \textbf{+4.1\%} absolute average accuracy boost over standard OPSD across diverse model scales and datasets. Further discussions confirm that AD-OPSD mitigates thinking collapse and generalizes robustly across different post-training alignment configurations, such as the student Non-Think setup.

In summary, our \textbf{contributions} are threefold:
\begin{itemize}
    \item We diagnose \textbf{Thinking Collapse} in reasoning OPSD through a multi-scale evidence chain: entropy-based gradient masking localizes the failure to high-student-entropy decision forks; token-level target analysis reveals severe Student-ET $\to$ Teacher-Non-ET discrepancies at initialization; and pointwise divergence mapping identifies high-risk suppression regions for real-time intervention.
    \item We propose \textbf{AD-OPSD}, a student-side, token-wise post-training framework that dynamically anchors high-suppression-risk tokens to a frozen base-model reference prior via an asymmetrical pointwise divergence gate, preserving native exploratory reasoning while retaining teacher corrective supervision.
    \item Extensive evaluations across competitive mathematical benchmarks and diverse model families show that AD-OPSD mitigates thinking collapse, recovers thinking density, and improves over standard OPSD by up to a $+4.1\%$ absolute average accuracy gain, with robust generalization to non-thinking alignment configurations.
\end{itemize}

\section{Thinking Collapse in OPSD}
\label{sec:thinking_collapse}
While On-Policy Self-Distillation (OPSD) has proven highly effective for standard alignment tasks, its application to reasoning models reveals a paradoxical performance degradation. Prior studies~\citep{kim2026does,kaur2026rethinking} trace the degradation to the suppression of epistemic verbalizations or fork tokens. Inspired by these pioneering findings, we introduce the quantitative metric of \textbf{thinking density} to represent the model's exploratory \textbf{thinking intensity} and investigate how this behavior changes across different self-distillation configurations.

\begin{figure}[t]
    \centering
    \includegraphics[width=0.9\linewidth]{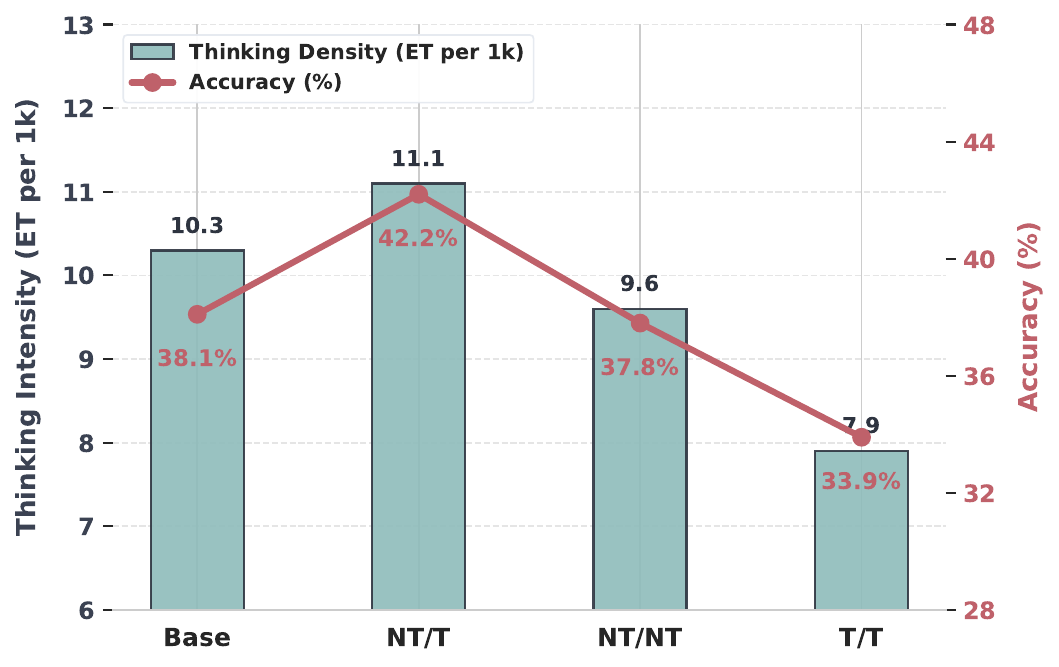}
    \caption{AIME25 Accuracy vs. Thinking Density (ET per 1k) Across OPSD Distillation Modes on Qwen3-1.7B. On-policy self-distillation under the student thinking mode (T/T) severely impairs reasoning performance, which is highly correlated with the sharp collapse of its thinking density.}
    \label{fig:et_phenomenon}
\end{figure}

Specifically, while prior literature often measures the absolute count of epistemic tokens (ET) within a generated rollout ($E(y) = \sum_{t} \text{count}(t, y)$), this raw count is highly sensitive to variations in sequence length. To establish a robust, length-invariant proxy for the model's underlying thinking intensity, we formally define \textbf{thinking density} as the frequency of epistemic tokens per 1,000 generated tokens:
\begin{equation}
    \text{Thinking Density} = \frac{\sum_{t \in \mathcal{T}} \text{count}(t, y)}{L(y)} \times 1000
\end{equation}
where $L(y)$ is the total sequence length, and $\mathcal{T}$ represents the set of 10 epistemic markers acting as indicators of planning, uncertainty externalization, and backtracking~\citep{kim2026does}:
\begin{equation}
\begin{aligned}
    \mathcal{T} = \{&
    \text{wait}, \text{hmm}, \text{perhaps}, \text{maybe}, \text{actually}, \\
    &\text{alternatively}, \text{seems}, \text{might}, \text{likely}, \text{check}
    \}
\end{aligned}
\end{equation}
To empirically verify the impact of self-distillation on thinking density, we train Qwen3-1.7B across distinct OPSD configurations on the AIME25 benchmark, including Think Student / Think Teacher (\textbf{T/T}) and Non-Think Student / Think Teacher (\textbf{NT/T}) setups. The empirical results, illustrated in Figure~\ref{fig:et_phenomenon}, reveal a severe optimization pathology. While standard OPSD (T/T) severely suppresses the student's thinking density---collapsing it precipitously from the base baseline of 10.3 down to \textbf{7.9 ET per 1k}---it simultaneously causes a downstream accuracy decay from 38.1\% to \textbf{33.9\%} (a -4.2\% degradation). Conversely, disabling student thinking during training (NT/T) preserves a healthy thinking density of \textbf{11.1} during evaluation, allowing accuracy to climb to \textbf{42.2\%} (+4.1\% absolute boost over Base). This tight correlation strongly suggests that OPSD's reasoning degradation is closely linked to the pathology of \textbf{Thinking Collapse}, which suppresses the model's exploratory, intermediate reasoning steps.

Although NT/T achieves the strongest accuracy in Figure~\ref{fig:et_phenomenon}, it introduces a cross-mode distillation factor: a non-thinking student learns from a thinking teacher. This makes it difficult to determine whether the observed gain comes from OPSD itself or from mode-level transfer. In contrast, T/T keeps the student and teacher in the same thinking mode, thereby providing a cleaner setting to diagnose how OPSD directly affects the model's own reasoning trajectories. We therefore focus our subsequent diagnostic analysis on T/T, as it reveals the core failure mode in which OPSD suppresses the model's intrinsic exploratory and backtracking behavior.

\section{Diagnosing the Causes of Thinking Collapse}
\label{sec:diagnosis}

While Section~\ref{sec:thinking_collapse} establishes the empirical existence of Thinking Collapse in reasoning OPSD, the underlying mechanics of \textit{where} and \textit{why} this suppression occurs remain unclear. To unpack the root causes of this pathology, we conduct a systematic, multi-scale diagnostic investigation across three analytical phases:
\begin{itemize}
    \item \textbf{Entropy-Based Gradient Masking (Section~\ref{sec:masking}):} We perform targeted token masking based on student entropy during backpropagation to spatially isolate and localize the main region of thinking collapse.
    \item \textbf{Token-Level Discrepancy Analysis (Section~\ref{sec:transition}):} We analyze static target distributions at initialization to investigate the microscopic target discrepancies over epistemic tokens at critical decision forks.
    \item \textbf{Pointwise Divergence Mapping (Section~\ref{sec:map}):} We investigate the statistical properties of pointwise KL divergence to examine its correlation with thinking collapse and validate its potential as a dynamic guidance indicator.
\end{itemize}
This diagnostic roadmap directly exposes the underlying optimization pathology of Thinking Collapse, providing the empirical foundation for our proposed Adaptive Dual-Perspective OPSD (AD-OPSD) framework.

\begin{figure*}[t]
    \centering
    \includegraphics[width=1.0\textwidth]{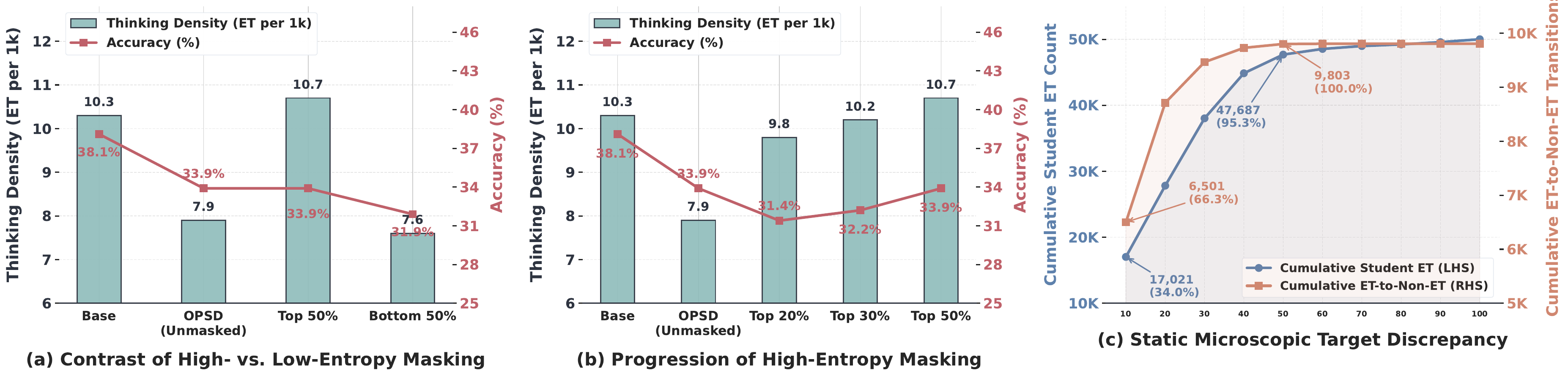}
    \caption{\textbf{Consolidated OPSD Diagnostics} under Token-Level Masking and Microscopic Target Discrepancy. Subfigures (a) and (b) illustrate OPSD performance under entropy-based masking (LHS: Thinking Density, RHS: Accuracy \%). Subfigure (c) depicts the cumulative student ET count (LHS, blue curve) and cumulative ET-to-Non-ET target discrepancy transitions (RHS, orange curve) across student entropy percentiles sorted in descending order at initialization (step 0). Both masking plots include Base and standard OPSD (Unmasked) as reference baselines.}
    \label{fig:masking_diagnostics}
\end{figure*}

\subsection{Localizing Thinking Collapse via Entropy-Based Gradient Masking}
\label{sec:masking}
To investigate where and how thinking collapse occurs in the generation sequence, we analyze how OPSD performance and thinking density respond to targeted token masking. In reasoning models, only a small fraction of tokens exhibit high entropy, acting as critical decision forks that steer the model toward diverse reasoning pathways~\citep{wang2026beyond}. Inspired by this, we study OPSD dynamics from the perspective of student entropy $H(y_i \mid y_{<i})$. Using the standard OPSD training settings~\citep{zhao2026self}, we partition the student's predictive token space dynamically during backpropagation and selectively mask out the self-distillation gradients under two polar configurations: (1) \textbf{Top-$p\%$ Student Entropy (SE) Mask} (where $p \in \{20, 30, 50\}$), which shields the student's most uncertain decision points, and (2) \textbf{Bottom-$50\%$ Student Entropy (SE) Mask}, which restricts distillation exclusively to these high-uncertainty regions. The experimental results, illustrated in Figures~\ref{fig:masking_diagnostics}(a) and \ref{fig:masking_diagnostics}(b), provide a systematic, multi-scale localization of thinking collapse, while exposing a critical learning trade-off. 

\paragraph{High-Entropy Supervision as the Primary Source of Collapse.}
The macro-level contrast of high-entropy versus low-entropy masking in Figure~\ref{fig:masking_diagnostics}(a) highlights the primary driver of this pathology. When applying the Bottom-50\% SE mask (distilling exclusively on high-entropy tokens), the evaluation thinking density collapses heavily to \textbf{7.6 ET per 1k}—even lower than the standard unmasked OPSD baseline of 7.9. Conversely, completely shielding the high-entropy region via the Top-50\% SE mask fully restores thinking density to \textbf{10.7 ET per 1k}, exceeding the pre-trained base baseline (10.3). This stark contrast strongly indicates that thinking collapse is primarily driven by the teacher's aggressive supervision over the student's highest-entropy states.

\paragraph{Suppressive Pressure Concentrated at Critical Decision Forks.}
A closer examination of the high-entropy region's progression—increasing the masked fraction from Top 20\% to Top 50\% in Figure~\ref{fig:masking_diagnostics}(b)—reveals the localized concentration of this suppressive force. Shielding only the Top-20\% SE tokens is highly sufficient to recover thinking density to \textbf{9.8 ET per 1k}, representing a massive rebound from standard OPSD (7.9). However, further widening the mask to Top-30\% and Top-50\% yields heavily diminishing returns, restoring density to \textbf{10.2} and \textbf{10.7} respectively. This decelerating growth curve demonstrates that suppressive pressure is highly concentrated within the very tip of the highest-entropy states (the top 20\% critical decision forks). Crucially, as we gradually recover more of the model's native thinking density (from 9.8 to 10.7 ET per 1k), downstream accuracy exhibits a corresponding climb from \textbf{31.4\%} to \textbf{33.9\%}, indicating a clear positive association between preserving native intermediate thinking steps and downstream performance.

\paragraph{Optimization Deadlock of Binary Gradient Masking.}
Despite these large differences in thinking density, a striking aspect of the results is that standard OPSD (\textbf{33.9\%}), Mask Top-50\% (\textbf{33.9\%}), and Mask Bottom-50\% (\textbf{31.9\%}) achieve relatively similar and degraded downstream accuracies. This performance deadlock arises from two distinct and opposite failure modes. While standard OPSD receives the teacher's full corrective signals but suffers from severe thinking collapse (Thinking Density = 7.9), the Mask Top-50\% setup preserves the model's native thinking capacity (Thinking Density = 10.7) but is deprived of the teacher's vital corrective feedback at critical decision forks. Meanwhile, the Mask Bottom-50\% configuration suffers from collapsed thinking density (7.6) while also being deprived of teacher gradients on low-entropy factual tokens. This detailed diagnosis suggests that simple binary gradient masking is caught in an optimization deadlock: we cannot resolve thinking collapse without sacrificing teacher correction, and vice-versa, highlighting the need for a non-binary, adaptive soft-gating formulation to balance thinking preservation and factual correction.

\subsection{Dissecting Thinking Collapse via Microscopic Transition Analysis}
\label{sec:transition}

To gain a deep, token-level understanding of how thinking collapse is triggered under standard OPSD, we conduct a microscopic transition analysis. Our diagnostic is a \textit{static target discrepancy analysis} conducted at the very beginning of training (using the initial checkpoints before any gradient updates), which isolates the raw, unmodified supervisory signals that dictate the initial gradient directions. We randomly sample 200 tasks from the OpenThought dataset and generate 4 independent rollouts under Qwen3-1.7B's thinking mode, yielding a total of 800 rollouts (3.9M tokens) scored by a ground-truth-conditioned teacher model at initialization.

The initial static distributions, illustrated in Figure~\ref{fig:masking_diagnostics}(c), reveal two crucial findings explaining the root cause of thinking collapse:
\begin{itemize}
    \item \textbf{High-Entropy Concentration of Epistemic Tokens:}
    The student model's native epistemic tokens (ETs) are heavily concentrated within high student entropy regions. Specifically, the top 10\% highest-entropy tokens contain \textbf{17,021} ETs (\textbf{34.03\%} of all generated ETs), and the top 50\% highest-entropy tokens contain \textbf{47,687} ETs (\textbf{95.35\%} of all ETs). This suggests that the model's native reasoning and intermediate thinking steps are structurally localized at high-entropy decision "forks" where the student actively explores.
    
    \item \textbf{Severe Top-1 Discrepancy (Student ET $\to$ Teacher Non-ET):} 
    Within these identical high-entropy regions, there is a substantial target discrepancy where the student's top-1 predicted token is an ET, but the teacher's corresponding top-1 predicted token is a Non-ET. Out of 9,807 total ET-to-Non-ET discrepancy events across the entire 3.9M token corpus, \textbf{6,501 (66.29\%)} are concentrated in the top 10\% highest-entropy interval, and \textbf{9,803 (99.96\%)} occur within the top 50\% highest-entropy region.
\end{itemize}
These initial static discrepancies generate strong negative gradients on the student's exploratory paths at step 0. This optimization force aggressively suppresses the student's ET probabilities, explaining why standard OPSD rapidly triggers thinking collapse.

\subsection{Mapping the Epistemic Space of Thinking Collapse via Pointwise Divergence}
\label{sec:map}

To resolve the thinking collapse trap, we need a robust, real-time guidance indicator to identify where intermediate thinking steps are under high risk of being suppressed during on-policy self-distillation. On exploratory tokens (e.g., intermediate verification steps like ``Wait''), the student's probability is typically much larger than the teacher's, yielding a substantial log probability ratio of $(\log P_s(y_i) - \log P_t(y_i)) \gg 0$~\citep{shen2026anti,kim2026rebellious,yang2026self}. To prevent false positives where the student probability is near-zero but the teacher's is even smaller, we weight this log ratio by the student's probability, resulting in the pointwise KL divergence contribution:
\begin{equation}
    C_i = P_s(y_i \mid y_{<i}) \left( \log P_s(y_i \mid y_{<i}) - \log P_t(y_i \mid y_{<i}) \right)
\end{equation}
where $y_i$ is the actually sampled token in the rollout sequence. 

\begin{figure}[t]
    \centering
    \includegraphics[width=1.0\linewidth]{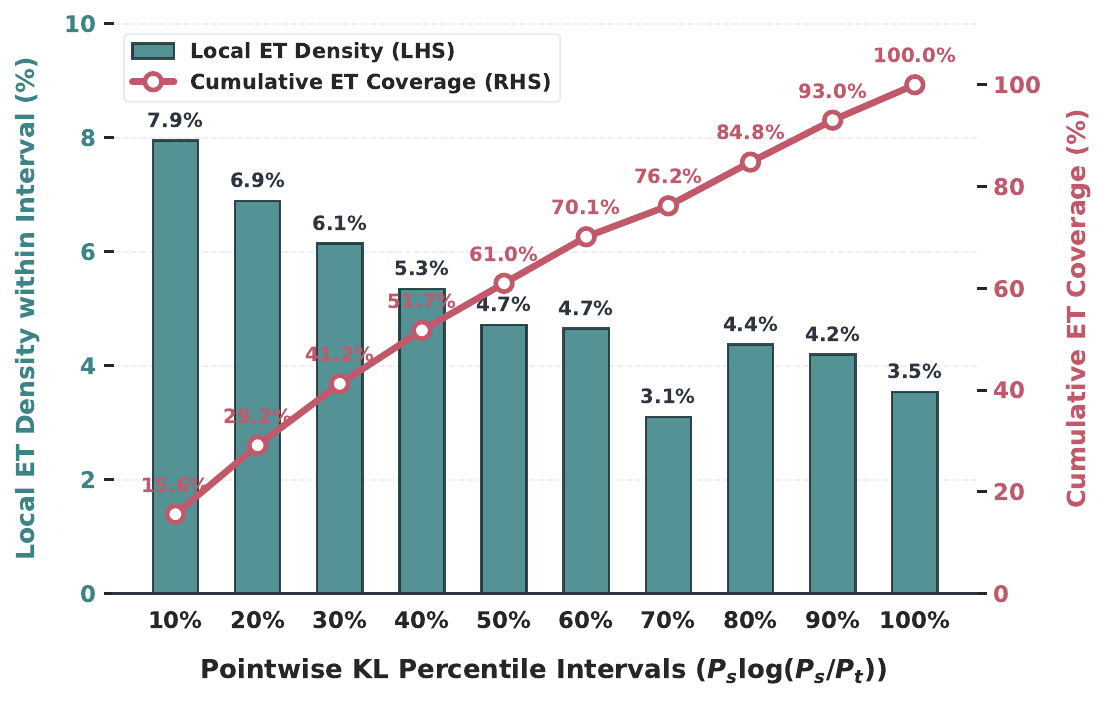}
    \caption{\textbf{Distribution and Concentration of Epistemic Tokens} across Pointwise KL Percentiles within High-Entropy Regions (Top 20\% Student Entropy) under Active Suppression ($P_s > P_t$). Pointwise KL serves as a highly robust guidance indicator of suppression risk, allowing AD-OPSD to protect vital thinking steps by shielding regions of high pointwise divergence.}
    \label{fig:et_by_kl_percentile}
\end{figure}

To validate whether the pointwise KL divergence contribution $C_i$ can serve as a robust proxy for locating these high-risk suppression regions, we focus exclusively on the active suppression region ($P_s > P_t$, i.e., $\log P_s - \log P_t > 0$) within the top 20\% student entropy. We sort these tokens by their pointwise KL $C_i$ in descending order and partition them into standard 10\% interval slices from 10\% to 100\%. The resulting local ET density and cumulative ET coverage are illustrated in Figure~\ref{fig:et_by_kl_percentile}.

The empirical distributions reveal that epistemic tokens (ETs) are exceptionally densely concentrated within regions of high pointwise divergence, which mathematically validates $C_i$ as a powerful real-time indicator of suppression risk:
\begin{itemize}
    \item \textbf{Dense ET Concentration in High Divergence Intervals:} As shown in Figure~\ref{fig:et_by_kl_percentile}, the local ET density is monotonically and sharply higher in the top pointwise KL intervals, representing a heavily suppressed thinking zone. Specifically, the local ET density peaks at \textbf{7.9\%} in the top 10\% highest interval, and steadily decreases to \textbf{6.9\%} (10\%--20\% interval), \textbf{6.1\%} (20\%--30\% interval), and down to under \textbf{3.5\%} at the lowest intervals. This stark contrast demonstrates that high pointwise divergence regions naturally cluster a high density of epistemic tokens.
    
    \item \textbf{High-Recall Cumulative Coverage:} Concurrently, the cumulative coverage curve exhibits a steep, near-linear rise at the beginning. Specifically, the top 10\% pointwise KL interval alone captures \textbf{15.6\%} of all generated ETs within this active suppression zone, the top 20\% highest intervals cover \textbf{29.2\%}, and the top 50\% highest intervals capture a substantial \textbf{61.0\%} of the total epistemic tokens, before eventually reaching \textbf{100.0\%} at the 100\% percentile.
\end{itemize}

This quantitative behavior provides a strong design justification for our adaptive framework as highlighted in our abstract and introduction. Since epistemic tokens are densely concentrated within high pointwise divergence regions under active suppression, local pointwise KL can serve as an exceptionally robust real-time indicator of suppression risk. By modulating the distillation objective using a soft-gating mechanism triggered by this pointwise divergence, we can dynamically assign larger weights to the reference prior in high suppression-risk regions. This allows us to successfully protect the model's native thinking steps while retaining OPSD's error-correcting power in other regions. This directly guides our AD-OPSD framework detailed in Section~\ref{sec:method}.

\section{Adaptive Dual-Perspective On-Policy Self-Distillation}
\label{sec:method}

To mitigate the thinking collapse, we propose \textbf{Adaptive Dual-Perspective On-Policy Self-Distillation (AD-OPSD)}, a robust control framework that dynamically moderates the self-distillation objective. Instead of binary token masking, which deprives the model of corrective teacher signals, AD-OPSD treats the pointwise divergence as a continuous indicator of thinking suppression risk. It dynamically moderates the teacher's influence and shifts trust to the frozen base model, preserving native reasoning capacity without losing OPSD's error-correcting power. AD-OPSD consists of two core components: (1) a dual-perspective target formulation, and (2) an adaptive gating mechanism.

\subsection{Dual-Perspective Target Formulation}
In standard OPSD, the student policy $\pi_\theta(y_i \mid y_{<i}, X)$ is optimized directly toward the ground-truth-conditioned teacher distribution $\pi_t(y_i \mid y_{<i}, X, \text{GT})$ via Forward Kullback-Leibler (KL) divergence. To protect the model's exploratory reasoning capacity from aggressive teacher gradients, we reformulate the target distribution for each token $y_i$ into a dynamically interpolated target $\pi^*_i$:
\begin{equation}
\label{eq:target}
\begin{aligned}
    \pi^*_i(v)
    ={}& U_i \pi_{\text{base}}(v \mid y_{<i}, X) \\
    &+ (1 - U_i) \pi_t(v \mid y_{<i}, X, \text{GT})
\end{aligned}
\end{equation}
where $\pi_{\text{base}}$ is the frozen base model acting as an \textit{epistemic anchor}, and $U_i \in [0, 1]$ represents the \textit{Teacher Unreliability Index}. Scoring student rollouts with the frozen base model $\pi_{\text{base}}$ anchors the target to the model's unsuppressed native reasoning prior, preventing policy degradation.

\subsection{Adaptive Gating Mechanism}
The unreliability index $U_i$ dynamically moderates the balance between the base prior and the corrective teacher. To safeguard exploration without compromising critical corrective signals, we propose a two-stage \textbf{Adaptive Gating Mechanism}:

\paragraph{Spatial Sandbox Localization.}
Grounded in our findings in Section~\ref{sec:masking}, we restrict the gating intervention strictly to the student's high-entropy region $\mathcal{V}_{\text{High-}H}$, defined as the top-$p\%$ student entropy. For any token outside this high-entropy sandbox ($y_j \notin \mathcal{V}_{\text{High-}H}$), the teacher's feedback is trusted completely, and we statically set $U_j = 0$, leaving standard factual error correction untouched.

\paragraph{Pointwise Divergence Gating.}
Within the high-entropy sandbox ($y_i \in \mathcal{V}_{\text{High-}H}$), we compute the unreliability index $U_i$ using an asymmetrical soft pointwise divergence gating mechanism. To selectively protect exploratory paths under suppression risk while preserving constructive teacher feedback, the gating is active exclusively in regions of positive pointwise log-ratio:
\begin{equation}
\small
    U_i = 
    \begin{cases} 
      \sigma \left( \frac{P_\theta(y_i \mid \cdot) \left(\log P_\theta(y_i \mid \cdot) - \log P_t(y_i \mid \cdot) \right)}{\tau} \right) & \text{if } u_i > 0 \\
      0 & \text{otherwise}
   \end{cases}
\end{equation}
where $\sigma(\cdot)$ is the standard sigmoid, $P_\theta(y_i \mid \cdot)$ is shorthand for $P_\theta(y_i \mid y_{<i}, X)$, $u_i$ is $\log \frac{P_\theta(y_i \mid \cdot)}{P_t(y_i \mid \cdot)}$, and $\tau$ is a temperature parameter (set to $1.0$ as default). In implementation, $U_i$ is computed from stop-gradient student probabilities and is used only to construct the token-level target distribution.

This asymmetrical formulation ensures that the epistemic anchor $\pi_{\text{base}}$ is blended in smoothly via the pointwise KL contribution only when the student probability on a sampled token exceeds the teacher's ($P_\theta > P_t$), representing potential thinking suppression. Otherwise, when the teacher endorses or corrects the student's output ($P_\theta \le P_t$), $U_i$ collapses to $0.0$, allowing corrective gradients to proceed at full strength.

\paragraph{Static Blending Variant.}
We also propose a simpler baseline variant named \textbf{Static Dual Perspective OPSD (SD-OPSD)}. This foundational variant fixes the unreliability index inside the active suppression region ($\log P_s > \log P_t$) of the sandbox to a constant factor, $U_i = \alpha$ (where $\alpha = 0.5$ as default). While this static variant can mitigate thinking collapse by anchoring the target, its uniform penalty does not distinguish between different levels of suppression risk across tokens. In contrast, our default \textbf{Dynamic Blending} formulation represents a further optimization that treats pointwise student-teacher divergence as a continuous indicator, selectively and proportionally shielding critical reasoning steps where $U_i$ is active ($U_i > 0$), while letting standard corrections where no suppression is detected ($U_i = 0$ exactly) proceed at full strength.

\subsection{AD-OPSD Loss Formulation}
The final optimization objective of AD-OPSD minimizes the KL divergence toward our reformulated target $\pi^*_i$ inside the high-entropy sandbox $\mathcal{V}_{\text{High-}H}$, while applying standard teacher-guided OPSD across the remaining tokens:
\begin{equation}
\scriptsize
\begin{aligned}
\mathcal{L}_{\mathrm{AD\text{-}OPSD}}(\theta)
={}&
\sum_{y_i \in \mathcal{V}_{\mathrm{High}\text{-}H}}
D_{\mathrm{KL}}\!\left(
\pi^*_i(\cdot)
\middle\|
\pi_\theta(\cdot \mid y_{<i}, X)
\right) \\
&+
\sum_{y_j \notin \mathcal{V}_{\mathrm{High}\text{-}H}}
D_{\mathrm{KL}}\!\left(
\pi_t(\cdot)
\middle\|
\pi_\theta(\cdot \mid y_{<j}, X)
\right).
\end{aligned}
\end{equation}

By optimizing this sequence-split objective, AD-OPSD mitigates the thinking collapse and the thinking-correction dilemma, preserving the model's native exploratory reasoning capacity while keeping factual error correction fully functional.

\section{Experimental Setup}
\label{sec:experiments}

\paragraph{Models and Datasets.} 
To verify the generalizability of our proposed framework, we evaluate AD-OPSD across two prominent reasoning model families at multiple scales: the \textbf{Qwen3}~\citep{yang2025qwen3} family (specifically Qwen3-1.7B and Qwen3-4B) and the widely used \textbf{DeepSeek-R1-Distill-Qwen-1.5B} model~\citep{guo2025deepseek}. For training data, we employ the mathematical reasoning subset of the \textbf{OpenThoughts}~\citep{guhaopenthoughts} dataset, sampling up to 3K high-quality problem-solution pairs equipped with chain-of-thought (CoT) reasoning trajectories. To comprehensively evaluate the models' competition-level mathematical reasoning and Test-Time Scaling (TTS) potential, we benchmark performance across four challenging mathematics benchmarks: \textbf{MATH}~\citep{lightman2023let}, \textbf{AIME 2024}~\citep{aime24}, \textbf{AIME 2025}~\citep{aime25}, and \textbf{HMMT 2025}~\citep{dekoninck2026matharena}. The details of the datasets are provided in Appendix~\ref{sec:dataset_specifications}.

\paragraph{Baselines.} 
To systematically demonstrate that our AD-OPSD framework mitigates the epistemic suppression bottleneck and preserves reasoning capabilities, we compare our approach against two foundational external baselines:
\begin{itemize}
    \item \textbf{Base:} The original pre-trained reasoning model evaluated under its native thinking mode, representing the model's native exploratory prior before alignment.
    \item \textbf{OPSD:} The standard On-Policy Self-Distillation~\citep{zhao2026self} baseline trained under the fully unmasked Student-Think/Teacher-Think paradigm, where both the active student and the ground-truth-guided teacher utilize their full thinking capabilities.
\end{itemize}
Additionally, we evaluate the \textbf{Static Blend ($w_i = 0.5$)} variant of our method (formulated in Section~\ref{sec:method}) as a direct internal ablation study to verify the necessity of our dynamic, compression-governed soft-gating mechanism.

\begin{table*}[t]
    \centering
    \caption{\textbf{Mathematical Reasoning Performance} (Accuracy, \%) Across Baselines and Models on Competition-Level Benchmarks. We compare our proposed AD-OPSD framework against the pre-trained Base Model, standard OPSD, and our static baseline variant SD-OPSD. Bold indicates the absolute best-performing model for each metric.}
    \label{tab:main_results}
    \resizebox{0.95\linewidth}{!}{
    \begin{tabular}{llccccc}
        \toprule
        \textbf{Models} & \textbf{Method} & \textbf{MATH} & \textbf{AIME 2024} & \textbf{AIME 2025} & \textbf{HMMT 2025} & \textbf{Average} \\
        \midrule
        \multirow{4}{*}{\textbf{Qwen3-4B}} 
        & Base & \textbf{95.8} & 73.6 & 65.6 & 41.9 & 69.2 \\
        & OPSD & 95.4 & 69.7 & 63.3 & 39.7 & 67.0 \\
        & SD-OPSD & 95.4 & 71.9 & 64.4 & 43.3 & 68.8 \\
        & \textbf{AD-OPSD (Ours)} & 95.4 & \textbf{74.4} & \textbf{66.4} & \textbf{44.7} & \textbf{70.2} \\
        \midrule
        \multirow{4}{*}{\textbf{Qwen3-1.7B}} 
        & Base & 90.4 & 48.6 & 38.1 & 22.2 & 49.8 \\
        & OPSD & 89.3 & 45.3 & 33.9 & 22.2 & 47.7 \\
        & SD-OPSD & 90.6 & \textbf{50.8} & 37.2 & \textbf{25.0} & 50.9 \\
        & \textbf{AD-OPSD (Ours)} & \textbf{91.0} & 50.6 & \textbf{40.6} & \textbf{25.0} & \textbf{51.8} \\
        \midrule
        \multirow{4}{*}{\textbf{DS-R1-Distill-1.5B}} 
        & Base & \textbf{83.3} & 24.7 & 21.4 & \textbf{13.1} & 35.6 \\
        & OPSD & 81.6 & 23.3 & 19.2 & 10.8 & 33.7 \\
        & SD-OPSD & 82.2 & \textbf{26.9} & 21.4 & \textbf{13.1} & \textbf{35.9} \\
        & \textbf{AD-OPSD (Ours)} & 82.0 & 26.1 & \textbf{23.3} & 11.7 & 35.8 \\
        \bottomrule
    \end{tabular}}
\end{table*}

\paragraph{Implementation Details.} 
Following the training protocol established in standard on-policy distillation frameworks~\citep{zhao2026self}, we fix the teacher policy as the initial, pre-trained starting checkpoint. This stabilized target setup acts as an implicit regularizer, preventing excessive policy deviation from the native exploratory prior. We perform full-vocabulary logit distillation across all training runs. All experiments are implemented using Parameter-Efficient Fine-Tuning via LoRA~\citep{hulora} and conducted on 4 NVIDIA RTX-5090 GPUs. Detailed hyperparameters, templates, and optimization configurations are provided in Appendix~\ref{sec:exp_details} and ~\ref{sec:prompt_templates}.


\section{Main Results}
\label{sec:main_results}
We evaluate the mathematical reasoning capabilities of our proposed \textbf{AD-OPSD} framework alongside our baselines across multiple competitive mathematics benchmarks. The main experimental results are summarized in Table~\ref{tab:main_results}. We compare our proposed adaptive framework (\textbf{AD-OPSD}) against the Base Model, standard OPSD, and our static baseline variant (\textbf{SD-OPSD}).

\paragraph{Consistent Performance Gains across Different Models and Datasets.}
Across all evaluated model scales and datasets, \textbf{AD-OPSD} consistently and systematically outperforms standard OPSD. While standard OPSD degrades reasoning performance compared to the Base baseline (e.g., plunging from 38.1\% to 33.9\% on AIME 2025 for Qwen-1.7B, and from 69.2\% to 67.0\% average for Qwen-4B), AD-OPSD completely mitigates this optimization pathology, delivering a substantial performance surge (e.g., achieving +4.1\% absolute improvement over standard OPSD on Qwen-1.7B average). Crucially, this performance gap is even more pronounced on \textbf{harder reasoning tasks}. For example, on the challenging AIME 2025 benchmark, AD-OPSD achieves massive absolute boosts of \textbf{+6.7\%} (40.6\% vs 33.9\%) on Qwen-1.7B and \textbf{+3.1\%} (66.4\% vs 63.3\%) on Qwen-4B over standard OPSD. This pattern indicates that preserving the student's exploratory reasoning steps is especially vital when dealing with complex, multi-step logical deductions where standard distillation gradients induce catastrophic thinking collapse.

\paragraph{Effectiveness of Static Anchoring and Superiority of Dynamic Gating.}
From the results in Table~\ref{tab:main_results}, we observe that even our proposed simpler static variant, \textbf{SD-OPSD}, is highly effective compared to standard OPSD. By anchoring high-entropy suppressed states to the unsuppressed base prior, SD-OPSD systematically outperforms OPSD across all backbones, delivering average accuracy boosts of up to \textbf{+3.2\%} absolutely over standard OPSD. This confirms that static anchoring within the sandbox is highly capable of safeguarding exploratory reasoning trajectories. However, compared to this static baseline, our default dynamic \textbf{AD-OPSD} framework achieves further superior performance on the Qwen models, reaching \textbf{70.2\%} vs \textbf{68.8\%} on Qwen-4B and \textbf{51.8\%} vs \textbf{50.9\%} on Qwen-1.7B. This result proves that, compared to a uniform blend, dynamically adjusting the unreliability index based on local pointwise KL is more effective, validating the necessity of our pointwise divergence soft-gating.

\section{Discussion and Ablation Studies}
\label{sec:discussion}

In this section, we conduct a series of empirical investigations to discuss and resolve the following four core questions regarding the performance, robustness, and parameters of AD-OPSD:
\begin{itemize}
    \item \textbf{Q1 (Thinking Density):} Does our proposed AD-OPSD successfully preserve and recover the student's thinking density under on-policy self-distillation?
    \item \textbf{Q2 (Generality):} Can our robust control framework generalize to non-thinking alignment paradigms (specifically the standard NT/NT distillation)?
    \item \textbf{Q3 (Context Horizon):} How does the maximum sequence length ($L_{\text{max}}$) used during training influence our method's behavior?
    \item \textbf{Q4 (Boundary and Weight):} What are the optimal sandbox spatial boundaries and gating weight functions to balance epistemic preservation and corrective feedback?
\end{itemize}

\subsection{Preservation of Thinking Density}
To verify whether our method can recover the thinking density under OPSD, we evaluate our proposed AD-OPSD framework against the Base model and standard OPSD (T/T) on Qwen3-1.7B. We track downstream reasoning accuracy and the corresponding thinking density (\textit{ET per 1k}) across the challenging AIME 2024 and AIME 2025 benchmarks.

\begin{table}[h]
    \centering
    \caption{\textbf{Reasoning Performance (Accuracy, \%) and Thinking Density (TD)} on Qwen3-1.7B across AIME benchmarks.}
    \label{tab:think_density}
    \resizebox{0.95\linewidth}{!}{
    \begin{tabular}{lcccc}
        \toprule
        & \multicolumn{2}{c}{\textbf{AIME 2024}} & \multicolumn{2}{c}{\textbf{AIME 2025}} \\
        \cmidrule(lr){2-3} \cmidrule(lr){4-5}
        \textbf{Method} & \textbf{Acc. (\%)} & \textbf{TD} & \textbf{Acc. (\%)} & \textbf{TD} \\
        \midrule
        Base & 48.6 & 10.3 & 38.1 & 10.6 \\
        OPSD & 45.3 & 7.8 & 33.9 & 7.9 \\
        \textbf{AD-OPSD} & \textbf{50.6} & \textbf{8.4} & \textbf{40.6} & \textbf{8.6} \\
        \bottomrule
    \end{tabular}}
\end{table}

The experimental results, summarized in Table~\ref{tab:think_density}, demonstrate several critical empirical advantages. First, standard OPSD triggers a severe collapse in thinking density, which plunges by \textbf{24.2\%} (from 10.3 to 7.8) on AIME 2024 and \textbf{25.5\%} (from 10.6 to 7.9) on AIME 2025. In contrast, AD-OPSD partially recovers the thinking density to \textbf{8.4} and \textbf{8.6} respectively. This recovery confirms that our asymmetrical unreliability gating shields the student's exploratory paths from destructive distillation gradients. Second, by preserving reasoning pathways while enforcing corrective teacher feedback, AD-OPSD improves accuracy over standard OPSD by \textbf{+5.3\%} on AIME 2024 and \textbf{+6.7\%} on AIME 2025. This validation shows that our dynamic target formulation can break the thinking-correction dilemma, enabling the model to retain more of its Test-Time Scaling potential while improving performance.

\begin{table}[h]
    \centering
    \small
    \setlength{\tabcolsep}{8pt}
    \caption{\textbf{Performance Comparison} (Accuracy, \%) of Qwen3-1.7B Trained Under NT/NT Alignment, Evaluated across NonThinking and Thinking Inference Modes.}
    \label{tab:nonthink_analysis}
    \begin{tabular}{lccc}
        \toprule
        \textbf{Method} & \textbf{AIME24} & \textbf{AIME25} & \textbf{Average} \\
        \midrule
        \multicolumn{4}{c}{\textit{NonThinking Inference Mode}} \\
        \midrule
        Base & 14.4 & 10.3 & 12.4 \\
        OPSD & 12.5 & 11.4 & 12.0 \\
        \textbf{AD-OPSD} & \textbf{14.4} & \textbf{12.5} & \textbf{13.5} \\
        \midrule
        \multicolumn{4}{c}{\textit{Thinking Inference Mode}} \\
        \midrule
        Base & 48.6 & 37.2 & 43.4 \\
        OPSD & 48.6 & 37.8 & 43.2 \\
        \textbf{AD-OPSD} & \textbf{51.9} & \textbf{44.2} & \textbf{48.1} \\
        \bottomrule
    \end{tabular}
\end{table}

\subsection{Generalization to Non-Thinking Alignment Paradigms}
To demonstrate the generalization of our method, we apply our robust control mechanism to the Non-Think Student / Non-Think Teacher (\textbf{NT/NT}) alignment configuration, where explicit intermediate reasoning tags are completely stripped during distillation. We train Qwen3-1.7B with standard OPSD and AD-OPSD strictly under the NT/NT setup, and evaluate the aligned models under both NonThinking and Thinking inference modes.

The empirical findings in Table~\ref{tab:nonthink_analysis} show that even in the absence of explicit intermediate reasoning chains, AD-OPSD mitigates phrasing suppression under NonThinking inference, raising the average accuracy from standard OPSD's 12.0\% to \textbf{13.5\%} (outperforming the Base baseline of 12.4\%). Furthermore, when evaluated under Thinking inference mode (activating test-time chain-of-thought), the performance boost is substantial. While standard OPSD achieves an average accuracy of only 43.2\%, AD-OPSD climbs to \textbf{48.1\%} (a \textbf{+4.9\% absolute improvement} over OPSD and \textbf{+4.7\%} over Base). This phenomenon suggests that standard OPSD can corrupt latent reasoning structures even under stripped NT/NT post-training, while AD-OPSD's protective anchoring helps safeguard these latent reasoning circuits and preserve test-time scaling capability.

\subsection{Influence of Maximum Training Context Length}
\begin{figure*}[t]
    \centering
    \includegraphics[width=1.0\linewidth]{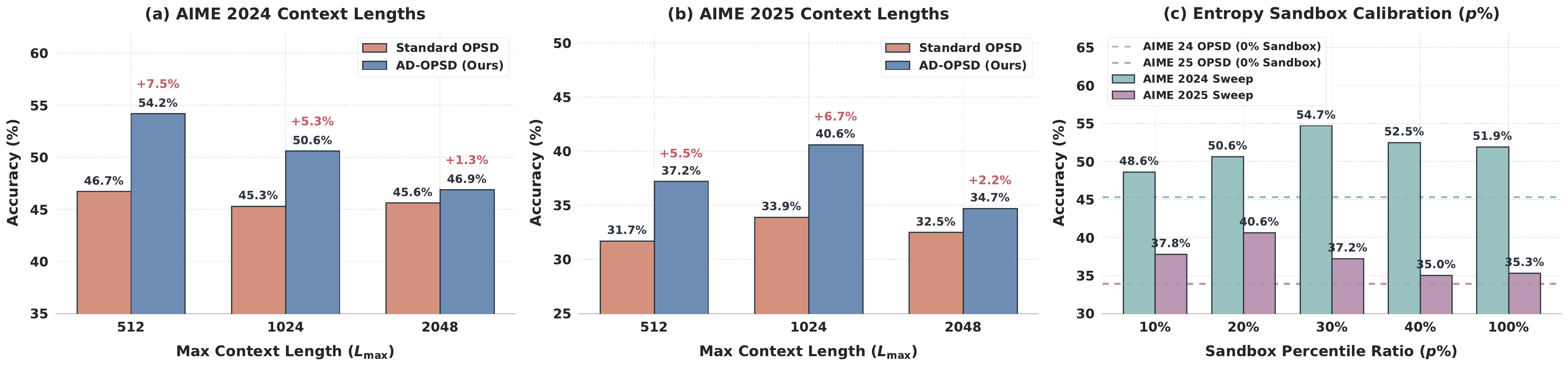}
    \caption{\textbf{Consolidated discussion and hyperparameter study} for Qwen3-1.7B on AIME benchmarks. Subfigures (a) and (b) present the downstream reasoning accuracy of Standard OPSD and our proposed AD-OPSD across varying maximum context lengths ($L_{\text{max}} \in \{512, 1024, 2048\}$) with absolute gains ($\Delta$) clearly annotated. Subfigure (c) presents the hyperparameter study on the Entropy Sandbox percentile ratio ($p\% \in \{10\%, 20\%, 30\%, 40\%, 100\%\}$) against standard OPSD (0\% Sandbox) represented as horizontal baselines.}
    \label{fig:hyperparam_study}
\end{figure*}

To analyze the influence of AD-OPSD to the maximum sequence length used during training ($L_{\text{max}}$), we train Qwen3-1.7B under three distinct context horizons: \textbf{512}, \textbf{1024}, and \textbf{2048} tokens, and plot the comparative accuracy and relative gains in Figures~\ref{fig:hyperparam_study}(a) and \ref{fig:hyperparam_study}(b).

The empirical findings, visualized in Figures~\ref{fig:hyperparam_study}(a) and \ref{fig:hyperparam_study}(b), reveal two key properties: AD-OPSD consistently and systematically outperforms standard OPSD across all context horizons. However, the margin of improvement ($\Delta$) exhibits a clear tapering trend on longer horizons. Specifically, the absolute gain on AIME 2024 narrows from \textbf{+7.5\%} at $L_{\text{max}}=512$ down to \textbf{+5.3\%} at $1024$, and further down to \textbf{+1.3\%} at $2048$ tokens. A similar trend is observed on AIME 2025, where the margin tapers from \textbf{+5.5\%} (512) and \textbf{+6.7\%} (1024) down to \textbf{+2.2\%} (2048).

This diminishing margin of improvement on longer context horizons is explained by a natural optimization convergence of target distributions:
\begin{equation}
\label{eq:len}
\begin{aligned}
    \lim_{i \to \infty}
    \Big\|
        &\pi_t(y_i \mid y_{<i}, X, \text{GT}) \\
        &- \pi_{\text{base}}(y_i \mid y_{<i}, X)
    \Big\|
    \to 0
\end{aligned}
\end{equation}
This behavior is strongly supported by recent literature. For instance, \citet{zhang2026fast} demonstrate that distillation signals are naturally concentrated within early reasoning prefixes, whereas \citet{ziheng2026less} identify that teacher supervision quality degrades on late tokens due to cumulative off-policy drift (termed \textit{Off-Policy Teacher Decay}). 

These insights align perfectly with our dual-perspective target formulation in Equation~\ref{eq:target}. During the early-to-mid stages of generation (short prefixes), prompt-response alignment constraints are highly active, and standard OPSD imposes aggressive information compression, making our protective anchoring highly effective. As the horizon extends, autoregressive uncertainty and off-policy drift cause the teacher's distribution to converge toward the base prior. Consequently, the active optimization conflict dissipates in long sequences, naturally narrowing the performance gap.

\subsection{Sensitivity and Ablation Analysis of the Sandbox Configuration}
To deeply investigate our design space, we perform two empirical analyses evaluating (1) spatial sensitivity to the sandbox percentile $p\%$, and (2) ablation of alternative gating weight functions. All experiments are conducted on Qwen3-1.7B across the competitive AIME benchmarks.

\paragraph{Spatial Sensitivity on Sandbox Percentile.} We perform a parameter sweep over $p\% \in \{10\%, 20\%, 30\%, 40\%, 100\%\}$ to calibrate the localized boundary of the high-entropy sandbox, establishing standard OPSD ($0\%$ Sandbox) as a reference baseline in Figure~\ref{fig:hyperparam_study}(c). The empirical results reveal that AD-OPSD consistently outperforms the standard OPSD baseline across all protection percentiles on both benchmarks. Notably, even under the extreme $100\%$ full-protection sandbox, AD-OPSD achieves \textbf{51.9\%} on AIME 2024 and \textbf{35.3\%} on AIME 2025 (exceeding standard OPSD's 45.3\% and 33.9\%, respectively), highlighting the effectiveness of our dual-perspective anchoring. Furthermore, downstream accuracy exhibits a distinct bell-curve trend as the sandbox size $p\%$ increases, supporting our spatial diagnostics in Section~\ref{sec:masking}. Restricting the anchoring shield to highly localized decision forks (the top 20\%--30\% predictive entropy) is sufficient to restore reasoning pathways, as AIME 2024 peaks at a wider sandbox of \textbf{30\%} (\textbf{54.7\%}) while the more challenging AIME 2025 peaks at a more compact sandbox of \textbf{20\%} (\textbf{40.6\%}). Beyond these thresholds, over-expanding the sandbox introduces a correction dilution problem that weakens vital teacher corrective gradients. Thus, localizing the protection to the top 20\%--30\% highest-entropy region strikes the optimal balance, and we select $p\% = 20\%$ as default.

\paragraph{Ablation on Gating Weight Functions.} Within the active suppression region ($\log P_s > \log P_t$) of the top-20\% Entropy Sandbox, we ablate our default thresholded pointwise KL sigmoid gate against three alternative weighting functions: (1) a pure log-ratio gate (\textit{Pure Log-Ratio}, $\sigma(\log P_s - \log P_t)$), (2) a global vocabulary-level uncertainty difference gate (\textit{Global Entropy}, $\sigma(H_s - H_t)$), and (3) a hyperbolic tangent mapping of the pointwise KL (\textit{Tanh Pointwise}, $\tanh(P_s(\log P_s - \log P_t))$). 

\begin{table}[h]
    \centering
    \small
    \caption{\textbf{Ablation Study} on Alternative Gating Formulations on Qwen3-1.7B.}
    \label{tab:gating_ablation}
    \resizebox{1.0\linewidth}{!}{%
    \begin{tabular}{lccc}
        \toprule
        \textbf{Gating Weight Function} & \textbf{AIME 24} & \textbf{AIME 25} & \textbf{Average} \\
        \midrule
        Base Model & 48.6 & 38.1 & 43.4 \\
        Standard OPSD & 45.3 & 33.9 & 39.6 \\
        \midrule
        Pure Log-Ratio & \textbf{51.9} & 35.3 & 43.6 \\
        Global Entropy & 50.3 & 36.7 & 43.5 \\
        Tanh Pointwise  & 51.1 & 35.3 & 43.2 \\
        \midrule
        \textbf{Sigmoid Pointwise} & 50.6 & \textbf{40.6} & \textbf{45.6} \\
        \bottomrule
    \end{tabular}}
\end{table}

The empirical results in Table~\ref{tab:gating_ablation} demonstrate that our proposed \textbf{Sigmoid Pointwise} formulation achieves the highest average reasoning accuracy (\textbf{45.6\%}) and peaks at \textbf{40.6\%} on the challenging AIME 2025 benchmark, outperforming the next-best alternative by \textbf{+3.9\% absolute}. In contrast, while the \textit{Pure Log-Ratio} gate excels on the simpler AIME 2024, its performance sharply decays to \textbf{35.3\%} on AIME 2025, revealing that lacking student-confidence dampening ($P_s$) leaves it vulnerable to false-positive protection from near-zero probability noise. Furthermore, the \textit{Global Entropy} gate yields stable but sub-optimal performance (\textbf{43.5\%} average) due to its lack of token-specific granularity, while the \textit{Tanh Pointwise} gate (\textbf{43.2\%} average) suffers from steep gradient transitions that destabilize early optimization. Ultimately, this ablation justifies our thresholded pointwise KL sigmoid gating as the most robust and precise mechanism for protecting exploratory reasoning pathways.

\section{Related Work}
\label{sec:related_work}

Our work is closely related to three lines of research: reasoning post-training and test-time scaling, on-policy distillation and its pathologies, and token-level intervention for preserving reasoning behavior.

\subsection{Reasoning Post-Training and Test-Time Scaling}
Exposing and sampling intermediate reasoning paths (e.g., chain-of-thought) lies at the core of unleashing LLM reasoning capabilities~\citep{wei2022chain,wang2023selfconsistency}. While outcome or step-level supervision provides dense feedback to guide these paths~\citep{cobbe2021training,lightman2023let}, process-level alignment studies further suggest that optimizing only final outcomes can damage reasoning integrity even when final accuracy improves~\citep{wu2025reasonkepp}. Recent advances in reasoning models demonstrate that enabling longer rollouts and test-time scaling compute is crucial to high performance on complex reasoning tasks~\citep{snell2024scaling,guo2025deepseek}. At the same time, recent analyses caution that reasoning quality is not determined by length alone: redundant self-doubt can induce overthinking, self-improvement training can suffer from collapse on complex reasoning tasks, and efficiency-oriented objectives can favor denser reasoning traces without sacrificing accuracy~\citep{peng2025overthinking,zhong2026better,peng2026thinkdense}. These works primarily study how to elicit, allocate, or compress reasoning compute. However, they do not address why standard post-training optimization objectives might inadvertently erase these vital intermediate paths. This optimization pathology motivates us to investigate the underlying causes of thinking collapse under on-policy self-distillation, shifting the focus from increasing generation length to protecting native thinking density.

\subsection{On-Policy Distillation and its Pathologies}
To bridge the train-test mismatch of standard sequence-level distillation, on-policy distillation (OPD) optimizes the student policy over its own sampled trajectories~\citep{agarwal2024onpolicy}. Although OPD avoids exposure bias, recent studies highlight its fragility. For instance, successful alignment requires compatible student-teacher thinking patterns~\citep{li2026rethinkingopd}, failing which can lead to imbalanced token-level supervision, prefix-drift, or tokenizer mismatch~\citep{fu2026revisitingopd,zhu2026manyfaces}. To leverage privileged teacher information while avoiding these pathologies, on-policy self-distillation (OPSD) has emerged to condition the policy on gold solutions or detailed feedback~\citep{zhao2026self,he2026selfdistillationzero}. Different from general OPD/OPSD studies, we show that the core failure mode of OPSD on reasoning tasks is not global mismatch, but rather a localized collapse of high-student-entropy thinking regions, which we resolve through selective reference anchoring.

\subsection{Token-Level Intervention and Fork Suppression}
Our diagnosis of thinking collapse is closely related to prior investigations on self-distillation risks~\citep{kim2026does,kaur2026rethinking}. Specifically, \citet{kim2026does} show that distillation suppresses epistemic verbalization, while \citet{kaur2026rethinking} identify that privileged-context targets suppress high-entropy decision forks. Most closely concurrent with our work, Purified OPSD~\citep{shen2026purified} addresses this suppression pathology by decomposing the teacher update using reference-only prompting to filter reference-specific shortcuts. 
While highly relevant, our approach differs fundamentally in mechanism: whereas Purified OPSD relies on teacher-side target reconstruction across all tokens, AD-OPSD operates directly on the student side. By restricting intervention to a localized high-entropy sandbox and scaling the unreliability index via pointwise KL, our method provides a dynamic, uncertainty-aware control gate to protect critical reasoning paths. Our formulation is also related in spirit to token-level self-evolution training, which dynamically regularizes uncertain learning states at a finer granularity than sequence-level objectives~\citep{peng-etal-2023-token}. This selective student-side intervention aligns perfectly with findings that reasoning updates concentrate heavily on high-entropy decision tokens~\citep{wang2026beyond,xu2026tip}, offering a complementary and highly precise control framework.

\section{Conclusion}
\label{sec:conclusion}

In this paper, we diagnosed and mitigated \textbf{Thinking Collapse}, an optimization pathology in reasoning-oriented On-Policy Self-Distillation (OPSD) where the model's native intermediate reasoning behavior sharply declines. Through entropy-based gradient masking, token-level discrepancy analysis, and pointwise divergence mapping, we showed that this collapse is concentrated at high-student-entropy decision forks, where teacher gradients frequently suppress student epistemic tokens into non-epistemic targets.

To address this pathology, we proposed \textbf{Adaptive Dual-Perspective OPSD (AD-OPSD)}, a robust post-training framework that dynamically interpolates the teacher distribution with a frozen base-model reference prior. By applying an asymmetrical pointwise divergence gate within a localized high-entropy sandbox, AD-OPSD preserves native thinking behavior while retaining OPSD's corrective supervision. Experiments across competitive mathematical reasoning benchmarks demonstrate that AD-OPSD consistently improves over standard OPSD, yielding up to a \textbf{+4.1\%} absolute average accuracy gain, recovering thinking density, and generalizing to non-thinking alignment configurations. These findings suggest that stable reasoning post-training requires not only stronger supervision, but also selective protection of the model's native exploratory reasoning paths.

\section*{Limitations}
\label{sec:limitations}
Despite the consistent empirical gains and robust theoretical justifications of our proposed framework, several limitations warrant further research:
\begin{itemize}
    \item \textbf{Model Scale and Parameter-Efficient Training:} Due to computational resource constraints, our empirical evaluation is primarily restricted to compact model scales trained using Parameter-Efficient Fine-Tuning via LoRA. Although our results demonstrate clear statistical significance and systematically mitigate thinking collapse at these scales, examining the scaling laws and generalizability of AD-OPSD on ultra-large models (e.g., 32B, 70B, or larger frontier models) and under full-parameter fine-tuning is a vital direction for industrial-scale deployment.
    \item \textbf{Domain Specialization in Mathematics:} Our diagnostics and main evaluations are strictly situated within competition-level math reasoning benchmarks. Mathematics is selected because symbolic deduction naturally exhibits dense, long-horizon chain-of-thought trajectories where thinking collapse is exceptionally pronounced. However, the generalizability of our dual-perspective anchoring framework to non-mathematical complex reasoning domains (such as symbolic logic, code generation, or multi-hop reading comprehension) remains to be extensively explored.
\end{itemize}


\bibliography{custom}

\appendix
\section{Dataset Specifications}
\label{sec:dataset_specifications}

To support reproducibility, we detail the training corpus, downstream benchmarks, optimization settings, method-specific hyperparameters, evaluation protocol, and prompt templates used in our experiments.

\subsection{Training Corpus}
We sample up to 3K high-quality mathematical reasoning problem-solution pairs from the open-source OpenThoughts dataset~\citep{guhaopenthoughts}. Each training example contains a problem statement and a reference solution used as privileged teacher context during self-distillation. The same sampled training subset is used across all trained OPSD variants, including standard OPSD, SD-OPSD, and AD-OPSD, so that observed differences arise from the distillation objective rather than from data variation. Downstream benchmark problems are used only for evaluation.

OpenThoughts is well suited for our study because its examples contain dense, multi-step chain-of-thought (CoT) reasoning trajectories. Such trajectories naturally include high-entropy decision forks, self-verification steps, and backtracking behavior, making the dataset an appropriate setting for diagnosing whether OPSD suppresses native exploratory reasoning.

\subsection{Downstream Evaluation Benchmarks}
We benchmark the mathematical reasoning capacity and test-time scaling behavior of aligned models across four competition-grade math datasets:
\begin{itemize}
    \item \textbf{MATH / MATH500~\citep{lightman2023let}:} We evaluate on the commonly used MATH500 subset, which covers high-school competition-level mathematical problems across algebra, number theory, geometry, probability, combinatorics, prealgebra, and precalculus.
    \item \textbf{AIME 2024~\citep{aime24} and AIME 2025~\citep{aime25}:} The American Invitational Mathematics Examination contains challenging problems with integer answers between 0 and 999. These benchmarks require long-horizon symbolic reasoning and extensive verification, making them central for evaluating preservation of native thinking capacity.
    \item \textbf{HMMT 2025~\citep{dekoninck2026matharena}:} The Harvard-MIT Mathematics Tournament contains difficult competition problems that require rigorous mathematical precision and creative exploratory paths, posing a challenging testbed for distilled reasoning backbones.
\end{itemize}

\section{Experimental Details}
\label{sec:exp_details}
We provide the shared training configuration in Table~\ref{tab:training_config}, the method-specific AD-OPSD configuration in Table~\ref{tab:ad_opsd_config}, and the evaluation configuration in Table~\ref{tab:eval-params}. Unless otherwise stated for ablation studies, all OPSD variants use the same training data, LoRA configuration, optimizer, batch size, number of training steps, and sampling settings.

\begin{table}[h]
\centering
\caption{Shared Training Configuration for OPSD Variants}
\label{tab:training_config}
\resizebox{\linewidth}{!}{
\begin{tabular}{lc}
\toprule
\textbf{Parameter} & \textbf{Value} \\
\midrule
Learning Rate & $5 \times 10^{-6}$ \\
Effective Batch Size & 16 \\
\midrule
LoRA Rank ($r$) & 64 \\
LoRA Alpha ($\alpha$) & 128 \\
LoRA Target Modules & 
\begin{tabular}[c]{@{}c@{}}
gate\_proj, up\_proj, \\
down\_proj
\end{tabular} \\
\midrule
Max Completion Length & 1024 \\
\midrule
Number of Generations per Prompt & 1 \\
Sampling Temperature & 1.1 \\
Training Steps & 188 \\
\midrule
Optimizer & AdamW \\
Precision & bfloat16 \\
Logit Distillation & Full vocabulary \\
\bottomrule
\end{tabular}}
\end{table}

\begin{table}[h]
\centering
\caption{Default AD-OPSD and SD-OPSD Method Configuration.}
\label{tab:ad_opsd_config}
\resizebox{\linewidth}{!}{%
\begin{tabular}{ll}
\toprule
\textbf{Parameter} & \textbf{Value} \\
\midrule
Reference Prior & Frozen initial/base model \\
Teacher Policy & Frozen initial checkpoint with ground-truth context \\
Entropy Sandbox & Top-$20\%$ student-entropy tokens by default \\
Gate Activation Region & Active suppression region ($\log P_s > \log P_t$) \\
Gate Score & $P_s(y_i)(\log P_s(y_i)-\log P_t(y_i))$ \\
Gate Temperature $\tau$ & 1.0 \\
Gate Gradient & Stop-gradient student probabilities \\
DS-OPSD Static Weight & $\alpha=0.5$ inside the entropy sandbox \\
\bottomrule
\end{tabular}}
\end{table}

For the sandbox sensitivity study in Section~\ref{sec:discussion}, we sweep $p\% \in \{10\%, 20\%, 30\%, 40\%, 100\%\}$ while keeping the other training settings fixed. For the context-length study, we vary the maximum completion length over $\{512, 1024, 2048\}$ tokens. For the gating ablation, we replace only the gating weight function and keep the top-$20\%$ entropy sandbox fixed.

\begin{table}[h]
\centering
\caption{Evaluation Parameters.}
\label{tab:eval-params}
\begin{tabular}{ll}
\toprule
\textbf{Parameter} & \textbf{Value} \\
\midrule
Max New Tokens & 38912 \\
Thinking Mode & \begin{tabular}[c]{@{}l@{}}Enabled unless evaluating\\NonThinking mode\end{tabular} \\
Top-p & 0.95 \\
Top-k & -1 \\
Min-p & 0.0 \\
Presence Penalty & 0.0 \\
Samples per Prompt & 4 / 12 \\ 
Temperature & 1.0 \\
\bottomrule
\end{tabular}
\end{table}

During evaluation, we sample 4 completions per prompt for MATH500 and 12 completions per prompt for AIME 2024, AIME 2025, and HMMT 2025, and report the mean accuracy over sampled completions following~\citet{zhao2026self}. All experiments were conducted using 4 RTX5090 GPUs with gradient checkpointing and Flash Attention 2~\citep{dao2023flashattention2} for memory efficiency.

\subsection{Answer Extraction and Scoring}
\label{sec:answer_extraction}

All models are prompted to put the final answer inside \verb|\boxed{}|. We evaluate mathematical correctness by exact match after normalized answer extraction. When a response contains one or more boxed answers, we extract the final boxed expression as the model's answer. If no boxed expression is present, we fall back to the final explicit answer-like expression in the completion. We normalize common formatting artifacts such as whitespace, surrounding punctuation, and simple LaTeX wrappers before comparison. For AIME-style benchmarks, answers are compared as normalized integer strings in the expected $0$--$999$ range. For MATH500 and HMMT-style problems, we compare the normalized extracted expression against the benchmark reference answer.

\subsection{Thinking and Non-Thinking Modes}
\label{sec:thinking_modes}

Our main OPSD experiments use the thinking mode, where the student is prompted to reason step by step and output a boxed final answer. In NonThinking configurations (e.g., NT/NT analysis), explicit intermediate reasoning instructions are removed and the model is prompted to provide the final answer directly. For Thinking inference, we use the step-by-step prompt template in Section~\ref{sec:prompt_templates}; for NonThinking inference, we use the same problem format but omit the step-by-step instruction. This separation allows us to test whether post-training damages only explicit reasoning traces or also the latent reasoning behavior recoverable under Thinking inference.

\definecolor{RoyalBlue}{RGB}{65, 105, 225}
\definecolor{OliveGreen}{RGB}{85, 107, 47}

\providecommand{\promptvar}[1]{\texttt{\textcolor{RoyalBlue!70!black}{\{#1\}}}}
\providecommand{\promptcard}[3]{%
    \begin{center}
    \setlength{\fboxsep}{7pt}%
    \fcolorbox{#1!55!black}{#1!5}{%
        \begin{minipage}{0.92\linewidth}
        \raggedright
        \textbf{\textcolor{#1!55!black}{#2}}\par\vspace{4pt}
        \small #3
        \end{minipage}%
    }%
    \end{center}
}

\section{Prompt Templates}
\label{sec:prompt_templates}

We fill the following templates before applying each model's chat template. For standard OPSD training, the student receives only the problem, while the teacher receives the same problem together with the reference solution as privileged context. The reference prior used by AD-OPSD is the frozen base model scored on the same student prefix without ground-truth context.

\promptcard{RoyalBlue}{Student Prompt Template}{%
{\ttfamily
Problem: \promptvar{problem}\par\medskip
Please reason step by step, and put your final answer within \textbackslash{}boxed\{\}.
}
}

The teacher model is guided by a cognitive transition prompt to encourage independent reasoning and verification rather than direct copying of the reference solution.

\promptcard{OliveGreen}{Teacher Prompt Template}{%
{\ttfamily
Problem: \promptvar{problem}\par\medskip
Here is a reference solution to this problem:\par
=== Reference Solution Begin ===\par
\promptvar{solution}\par
=== Reference Solution End ===\par\medskip
After reading the reference solution above, make sure you truly understand the reasoning behind each step --- do not copy or paraphrase it. Now, using your own words and independent reasoning, derive the same final answer to the problem above. Think step by step, explore different approaches, and don't be afraid to backtrack or reconsider if something doesn't work out:\par\medskip
Please reason step by step, and put your final answer within \textbackslash{}boxed\{\}.
}
}
Here \promptvar{solution} denotes the reference solution used as teacher context during distillation.

\promptcard{RoyalBlue}{NonThinking Prompt Template}{%
{\ttfamily
Problem: \promptvar{problem}\par\medskip
Please provide the final answer within \textbackslash{}boxed\{\}.
}
}

\end{document}